\documentclass[letterpaper, 10 pt, conference]{ieeeconf}  % Comment this line out if you need a4paper

\IEEEoverridecommandlockouts                              % This command is only needed if 
\usepackage{amsmath,amsfonts}
\usepackage[noend]{algorithmic}
\usepackage{algorithm}
\usepackage{array}
\usepackage[caption=false,font=normalsize,labelfont=sf,textfont=sf]{subfig}
\usepackage{textcomp}
\usepackage{stfloats}
\usepackage{url}
\usepackage{verbatim}
\usepackage{graphicx}
\usepackage{cite}
\usepackage{xcolor}
\usepackage{hyperref}
\usepackage{multirow}
\usepackage{booktabs}
\usepackage{pdfpages}
\usepackage{graphicx}
\definecolor{mydarkblue}{rgb}{0,0.08,0.45}
\hypersetup{
    colorlinks=true,
    linkcolor=mydarkblue,
    citecolor=mydarkblue,
    filecolor=mydarkblue,
    urlcolor=mydarkblue,
    }

\begin{document}
\title{\LARGE \bf COMPOSER: S\underline{c}alable and R\underline{o}bust \underline{M}odular \underline{Po}licies for \underline{S}nak\underline{e} \underline{R}obots}

% \author{Yuyou Zhang\thanks{Carnegie Mellon University, Pittsburgh, PA, USA.{\tt\footnotesize yuyouz@cmu.andrew.edu}}}
\author{Yuyou Zhang$^{1}$, Yaru Niu$^{1}$, Xingyu Liu$^{1} $and Ding Zhao$^{1}$% <-this % stops a space
% \thanks{*This work was not supported by any organization}% <-this % stops a space
\thanks{$^{1}$Yuyou Zhang, Yaru Niu, Xingyu Liu and Ding Zhao are with the Department of Mechanical Engineering,
        Carnegie Mellon University, Pittsburgh, PA 15213, USA.
        {\tt\small \{yuyouz, yarun, xingyul3, dingzhao\}@andrew.cmu.edu}}%
% \thanks{$^{2}$Yaru, Carnegie Mellon University, Pittsburgh, PA, USA.
%         {\tt\small b.d.researcher@ieee.org}}%
}

\maketitle
\begin{abstract}
Snake robots have showcased remarkable compliance and adaptability in their interaction with environments, mirroring the traits of their natural counterparts.
While their hyper-redundant and high-dimensional characteristics add to this adaptability, they also pose great challenges to robot control. 
% Many works on snake robots have focused on addressing the control challenges brought by their hyper-redundant and flexible characteristics. 
Instead of perceiving the hyper-redundancy and flexibility of snake robots as mere challenges, there lies an unexplored potential in leveraging these traits to enhance robustness and generalizability at the control policy level.
We seek to develop a control policy that effectively breaks down the high dimensionality of snake robots while harnessing their redundancy. In this work, we consider the snake robot as a modular robot and formulate the control of the snake robot as a cooperative Multi-Agent Reinforcement Learning (MARL) problem. 
% Soft robot control is particularly challenging due to its deformable shape and highly non-linear dynamics. 
% We decompose the high dimensionality inherent in the snake robot with the proposed modular control policy. 
Each segment of the snake robot functions as an individual agent.
Specifically, we incorporate a self-attention mechanism to enhance the cooperative behavior between agents. A high-level imagination policy is proposed to provide additional rewards to guide the low-level control policy.  We validate the proposed method 
% s\underline{c}alable and r\underline{o}bust \underline{m}odular \underline{po}licies for \underline{s}nak\underline{e} \underline{r}obots
COMPOSER with five snake robot tasks, including goal reaching, wall climbing, shape formation, tube crossing, and block pushing. COMPOSER achieves the highest success rate across all tasks when compared to a centralized baseline and four modular policy baselines.
% including MAPPO, MAT, MAT\_dec, and SMP. 
Additionally, we show enhanced robustness against module corruption and significantly superior zero-shot generalizability in our proposed method.
The videos of this work are available on our project page: \textcolor{blue}{\href{https://sites.google.com/view/composer-snake/}{https://sites.google.com/view/composer-snake/}}.

\end{abstract}

% \begin{IEEEkeywords}
% Article submission, IEEE, IE
% \end{IEEEkeywords}

\section{Introduction}
Snake robots have been extensively studied over the last decade. Their flexible and adaptive characteristics and bioinspired structure have enabled compliant interaction with the environment, facilitating their usage across areas including medical applications ~\cite{runciman2019soft}~\cite{wang2017cable}, extreme ~\cite{li2021self} or confined ~\cite{tang2022pipeline} ~\cite{Whitman-2018-122399} environment exploration.
The natural advantage of snake robots to work in unstructured environments is based on their material properties and morphology of their bodies \cite{runciman2019soft} rather than control strategy. In this work, we aim to improve the performance of snake robots by introducing a control strategy that leverages the inherent modularity of continuum robots. This modularity can be viewed from three perspectives: high dimensionality, scalability, and redundancy.

% Reinforcement learning has achieved success in numerous robotics tasks and demonstrated ability in complex control strategies and long-term planning, but learning to control continuum robots and soft robots is understudied compared to traditional robots.

% """Soft robotics aims to equip robots for the unpredictable needs of such situations by endowing them with capabilities that are based not in control systems but in the material properties and morphology of their bodies (Figure 1)""""

\begin{figure} [h]
\begin{center}
\includegraphics[width=1.\linewidth, page=7, trim = 14cm 3.3cm 1.cm 2cm, clip]{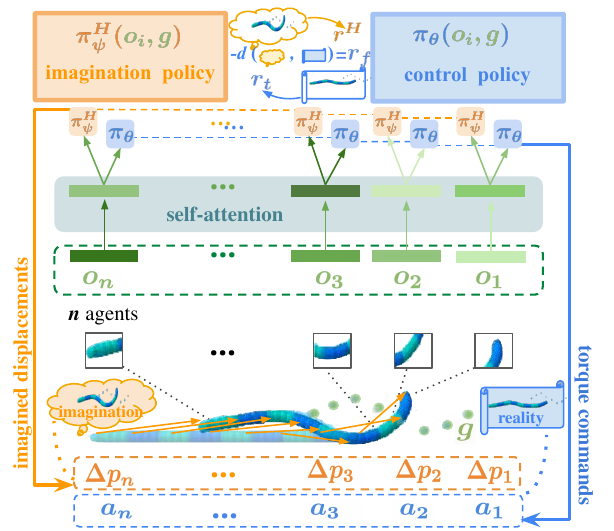}
\caption{Framework overview of COMPOSER. A snake robot with $n$ joints is formulated as $n$ agents. The modular control policy outputs individual torque commands, while the imagination policy forecasts an ideal displacement per step. The control policy is trained to both complete the task and adhere to the direction prescribed by the imagination policy.
}
\vspace{-10pt}
\label{fig: overview}
\vspace{-10pt}
\end{center}
\end{figure}

Snake robots, as continuum robots, possess a high-dimensional Degree of Freedom (DOF) and fall between the realms of soft robots and rigid-body robots.
% Snake robots are characterized by their snake-like morphology. 
This high-dimensional DOF complicates the system's kinematics and dynamics, posing a significant challenge to robot control. 
Model-based control methods ~\cite{kano2011decentralized,fu2020robotic,travers2018shape,rollinson2014torque} for snake robots suffer from this high dimensionality and usually struggle to approximate the dynamics model of the continuum body.
% ~\cite{zhu2022challenges}.  
Though reinforcement learning (RL) has shown remarkable model-free decision-making capabilities, 
learning an optimal policy in the presence of a high-dimensional space in snake robots can be costly~\cite{plaat2020deep}. Training a centralized policy, which requires exploring a high-dimensional space that grows exponentially with the snake robot DOF, can be highly inefficient.

% While learning-based methods have shown success in continuum robot control, collecting data for training either on real continuum robots or in a simulation with calibrated parameters is very expensive. 
% Most importantly, current learning-based methods lack the generalizability to match the scalability of the continuum robot morphology. And by assuming flawless functionality, current learning methods lack robustness against hardware failure which is allowed by continuum robot structural redundancy.

% As shown in \cite{graule2022somogym} Soft robots can be approximated by continuum robots. 
Snake robots have inherent modularity as continuum robots. This modularity enables scalability and flexibility, simplifying the snake robot design and facilitating easier maintenance. Structural scalability and flexibility empower snake robots to adapt to various tasks and unstructured environments~\cite {Whitman-2018-122399}. Current monolithic reinforcement learning ~\cite{chen2018hardware} ~\cite{jia2021coach} processes joint observations as inputs and produces joint actions as outputs. This centralized policy lacks generalizability to match the scalability of the snake robot. 

Modularity also gives snake robots hardware redundancy compared with traditional robots, which adds to system-level robustness against hardware failures, and allows for emergent behavior to adapt to perturbations~\cite{yim2003modular, itani2021motion, erkmen2002snake}. However, the presence of hardware redundancy becomes futile without an effective control policy to achieve policy-level robustness in snake robots. 
% By assuming flawless functionality, the monolithically learned policy lacks robustness against sensor or actuator failures, leading to the underutilization of the structural robustness provided by snake robot redundancy. 

To this end, our work is developed to break down the high dimensionality of snake robots while harnessing the structural scalability and redundancy brought by the inherent modularity. We first form the snake robot control problem as a multi-agent reinforcement learning problem and propose a shared modular torque control policy. Then we propose a high-level imagination policy to provide additional reward for control policy training in long-horizon tasks. We further utilize a self-attention mechanism to enhance inter-module communication. Five snake robot tasks are designed, showcasing manipulation and locomotion skills. Extensive experiments are conducted on five tasks to demonstrate the effectiveness of the proposed modular policy COMPOSER, in terms of success rate in normal functionality, robustness against hardware malfunction, and generalizability.
% The tasks are designed to showcase the advantages of snake robots compared with traditional robots, including adaptive and compliant interaction with the environment. 
% We first adopt a decentralized learning scheme to leverage the repetitive patterns inherent in snake robots and train a shared modular control policy. 
To the best of our knowledge, this is the first work that investigates the robustness and generalizability of a modular torque control policy for snake robots. The overview of our proposed method COMPOSER is shown in Fig. \ref{fig: overview}. 
% in the setting of decentralized modular policy learning for robot control.

The contributions are summarized as follows. 
\begin{itemize}
    \item We propose the first communication-enhanced decentralized modular control policy for snake robots to accommodate their inherent structural modularity.
    \item We introduce an imagination policy for efficient morphology-aware planning in long-horizon tasks.
    \item Our proposed method, COMPOSER, has demonstrated superior performance compared to strong baselines in all five snake robot tasks and showed better robustness and zero-shot generalizability.
\end{itemize}

% On the other side of this high dimensionality, the morphology of the snake robot shows repetitive patterns. Therefore , by leveraging a decentralized learning scheme, the adoption of modular policy appears to be an efficient way to reduce the dimension of the action space.

% Besides, the modular controller can utilize the structural similarities of all repetitive segments, enabling it to adapt to snake robots with varying joint numbers and, consequently, attain better generalizability. With this policy modularity and the structural modularity inherent in snake robots, we can have a robust policy to match the structural redundancy and a generalizable policy to match the structural scalability.

\section{Related Work}
\subsection{Snake and Continuum Robot Control}

% Snake robot as a continuum manipulator is a discretization of the soft robots, including pneumatic bellows~\cite{graule2022somogym,drotman2018application}, tendon-driven soft manipulator~\cite{xu2022adaptive,wang2016three,yip2014model}. Snake robots feature a finite number of actuators, applying forces/torques to the backbone at a preselected set of locations. 
% uncomment in soft robot workshop version!

% Continuum robot control has long been impeded by the complexity of the dynamics of continuous deformations. Various kinematics and dynamics models have enabled control of the continuum robots~\cite{wen2023modeling,camarillo2009configuration,xu2022adaptive,yip2014model}. ~\cite{xu2022adaptive}~\cite{yip2014model} use Jacobian to characterize the local linear relationship between joint observations and joint actions.
% and use a controller to output joint action that includes actions for each effective individual actuator within the robots
% These methods usually suffer to approximate the underlying dynamics. 
Early works of snake robots model-based control mainly focus on reproducing the innate behavior of real snakes. Various approaches have been explored, including generating torque proportional to the curvature derivative of the body curve~\cite{kano2011decentralized}, applying a serpenoid traveling wave propagated from the head to the tail~\cite{fu2020robotic}, utilizing gaits for shape based control~\cite{travers2018shape,rollinson2014torque}. However, simplifications made in these model-based approaches constrain the versatility of the modular snake robot, hurting compliance and flexible locomotion ability. 
% Reinforcement learning has been applied to fixed-base continuum manipulator tracking tasks\cite{satheeshbabu2019open,thuruthel2018model}. 
Many recent works in snake robots control have exhibited reduced reliance on the model, including co-optimization of morphology and control  \cite{sunco}, locomote with neural-network and CPG-based control\cite{liu2020learning}, coach-based reinforcement learning~\cite{jia2021coach}, reinforcement learning for snake robot tracking control\cite{bing2020perception}. However, ~\cite{sunco,jia2021coach,bing2020perception} use wheels to simplify snake robots and focus on in-plane control, leaving a gap in the exploration of complex tasks, such as manipulation and 3D locomotion in unstructured environments. Many prior studies have not fully leveraged the inherent modularity of snake robots. 
While Sartoretti et al. first explore snake robot modularity through the decentralized learning of shape-based locomotion parameters, its applicability is limited by reliance on a shape-based controller~\cite{sartoretti2019distributed}. In contrast, our approach in this paper directly learns a control policy to generate actuator torque, which enables learning intricate manipulation and locomotion skills.

% The Pseudo-Rigid Body (PRB) model is widely used for soft robot manipulators. In the PRB approach, the backbone along a subsegment is modeled as a series of rigid links connected by springs\cite{su2009pseudorigid,huang20193d}. This formulation represents the simulation of the snake robot discussed in this paper.  
% % The PRB 3R model shows potential to describe the large deformation of the flexible body and has high computational efficiency \cite{huang20193d}. 
% Policy trained with the simulated snake robot can be extended to soft robots by calibrating the relationship between actual control inputs (such as tendon length or pneumatic pressure) and simulated joint torques\cite{graule2022somogym}.
% uncomment in soft robot workshop version!

\subsection{Modular Policy for Decentralized Control}
Decentralized control is common in biological motor control in living creatures.  For example, a majority of neurons in the octopus are found in the arms, which can independently control basic motions without input from the brain, allowing fast responses. Using modular policy for decentralized control has demonstrated efficiency, robustness, and generalizability~\cite{schilling2021decentralized,pigozzi2022evolving,whitman2023learning,huang2020one}. Schilling et al. illustrate that within a decentralized control paradigm, reinforcement learning-based motor control not only attains high performance but also showcases enhanced robustness and improved generalizability~\cite{schilling2021decentralized}. 
%However, training a separate control policy for each module lacks efficiency and scalability ~\cite{schilling2021decentralized}. 
Pigozzi et al. employ evolutionary computation to optimize a shared neural controller for Voxel-based Soft Robots ~\cite{pigozzi2022evolving}. Whitman et al. use modular policy for reconfigurable robots~\cite{whitman2023learning}. In ~\cite{huang2020one}, a shared modular policy is trained with messages passing.  
%between modules as a pre-trained universal policy to control a variety of robot morphologies but requires a large set of robot variants for pretraining and is only validated on 2D MuJoCo environment. 
% Whether a shared modular policy trained on a snake robot can complete 3D tasks including rich interaction with the environment is unclear.
%Snake robots show inherent modularity. Their flexibility, compliance, and redundant characteristics all benefit from their modular structure.
~\cite{travers2018shape} shows that fewer individual degrees of freedom coupled means more efficient locomotion on irregular terrains. However, decentralized control has primarily been applied to snake robots within model-based control~\cite{kano2011decentralized,travers2018shape,sartoretti2019distributed}. As far as we know, this is the first work investigating learning a modular torque control policy to exploit the modularity of snake robots.

% Morphology-task graph ~\cite{furuta2022system}

% Structure-aware transformer policy \cite{hong2021structure}

% morphology (kinematic), material property(dynamics), or task level generalization $\to$ learn this prior \cite{gupta2022metamorph}

% Learning-based methods show promising results for handling complex underlying dynamics in deformable object manipulation~\cite{wang2022offline}.  

% Universal control policy are developed for skill level generalization\cite{nam2022skill}, task level generalization\cite{shi2022skill}, 
% generalization\cite{gupta2022metamorph}\cite{huang2020one}, or mitigating sim-to-real gap. 

\subsection{Multi-agent Reinforcement Learning}
% In MARL, decentralized execution is a more scalable approach compared with centralized learning which maps the joint observation to joint action~\cite{gupta2017cooperative}. 
On one hand, many works in MARL follow the Centralized Training Decentralized Execution (CTDE) framework, where decentralized policies are trained in a centralized fashion with extra information. MADDPG adopts actor-critic structures and learns a centralized critic~\cite{lowe2017multi}. Value-decomposition (VD) methods decompose the joint Q-function as a function of agents’ local Q-functions~\cite{sunehag2018value, rashid2020monotonic}. MAPPO demonstrates supreme performance in homogeneous multi-agent cooperative settings~\cite{yu2022surprising}. 
% For heterogeneous multi-agent cooperative settings, sharing a policy among agents potentially hurts the performance. 
HATPRO and HAPPO achieve promising results in heterogeneous settings, without parameter sharing among agents \cite{kuba2021trust}. 
On the other hand, some works employ centralized communication protocols during execution to share local information and augment the coordination among agents \cite{sukhbaatar2016learning, singh2018learning, niu2021multi}.
In this paper, we aim to utilize MARL to develop a modular policy tailored specifically for snake robots and continuum robots.

\section{Methodology}
% \subsection{Snake configuration}
% Our method is built on the following key observation. Snake robot as a continuum robot is characterized by their high dimensional state and action. On the other side of this high dimensionality, the mechanism of the snake robot shows repetitive patterns. Therefore, by leveraging a decentralized learning scheme, the adoption of modular policy appears to be an effective way to reduce the dimension of the action space and enable more efficient reinforcement learning. Besides, decentralized execution enhances robustness against hardware failures. Furthermore, the modular controller can utilize the structural similarities of all repetitive segments, enabling it to adapt to snake robots with varying joint numbers and, consequently, attain better generalizability. 
% By leveraging both policy modularity and the inherent structural modularity of snake robots, we can establish a resilient policy that aligns with structural redundancy, as well as a generalizable policy that corresponds to structural scalability.

We first introduce the Goal-conditioned CTDE paradigm for snake robot control as described in Section~\ref{subsec:ctde}. In Section~\ref{subsec:hp}, we elaborate on training a modular torque control policy with additional reward from an imagination policy.
%which predicts an imagined displacement.
Section~\ref{subsec: attention} introduces a self-attention mechanism to enhance efficient cooperation between agents. The full algorithm including the joint training of the modular torque control policy and the imagination policy is summarized in Algorithm \ref{algo:algo}. 

\subsection{Multi-agent Reinforcement Learning for Modular Policy}
\label{subsec:ctde}
In this paper, we formulate snake robot control as a fully cooperative MARL problem. 
Each actuator of the snake robot is considered an individual agent. 
We consider a multi-agent goal-conditioned partially observable Markov decision process for a snake robot with $n$ actuators, defined by $\langle \mathcal{S}, \mathcal{A}, O, R, P, \mathcal{G}, n, \gamma \rangle$.
$\mathcal{S}$ is the state space. 
$s \in \mathcal{S}$ describes states of all $n$ agents. 
Given the same configuration of all actuators with the snake robot, each agent shares an identical observation space and action space. 
Local observation for agent $i$ with $s$ is $o_i=O(s;i)$.  
At each time step, agent $i$ executes action $a_i$ sampled from a distribution generated by policy $\pi_\theta(.|o_i, g)$, which is parameterized by $\theta$ and shared by all agents. And all $n$ agents receive shared reward $r_1, \cdots, r_n=R(s, \boldsymbol{a}, g)$.
$R(s, \boldsymbol{a}, g)$ denotes the reward function and $g \in \mathcal{G}$ is a goal shared among all agents.
$\boldsymbol{a} = (a_{1}, a_{2}, \ldots, a_{n}) \in \mathcal{A}$ is the joint action of $n$ agents, which produces the next state $s^{\prime}$ with transition probability $P(s'|s,\boldsymbol{a})$.

We aim to learn a shared modular torque control policy $\pi_{\theta}$ for the snake robot. Each agent makes decisions individually while all agents execute actions simultaneously.  
The coupling between agents of the snake robot requires precise coordination among all cooperative agents to successfully complete tasks. The modular torque control policy $\pi_{\theta}$ is trained to jointly maximize the expected return $J(\theta)=\mathbb{E}_{\boldsymbol{a_t},s_t}\left[\sum_t \gamma^t R(s_t,\boldsymbol{a_t}, g)\right]$, with the discount factor $\gamma$. $\pi_{\theta}$ is optimized with the experiences of all agents simultaneously.
\subsection{Imagination-guided Control}
\label{subsec:hp}
In addition to using a shared torque control policy to output torque commands for each agent, we propose to use a shared modular imagination policy, $\pi^H_\psi$, parameterized by $\psi$ to predict per-step displacement for each agent. 
We use this imagination policy as an imaginary planner to decompose the long-horizon task into manageable steps and provide supplementary rewards for the torque control policy.

The imagination policy $\pi_\theta(.|o_i, g)$ is conditioned on the local observation $o_i$ and goal, and it generates distribution over anticipated displacement $\Delta p_i\in \mathbb{R}^3$. The displacement vector $\Delta p_i$ points from agent $k$'s current position $p_i\in \mathbb{R}^3$ to an imagined next-step position $p_i^{\prime}\in \mathbb{R}^3$, $p_i^{\prime}=p_i + \Delta p_i$. $\boldsymbol{\Delta p}$ denotes the predicted snake whole body shape transition.
Agents receive rewards by making good predictions so that the imagined next-step positions $p_i^{\prime}$ can be a reasonable subgoal which decomposes the long-horizon task. The reward function for imagination policy is defined to capture the proximity of the imagined next-step positions to the goal completion state, $R^H(\boldsymbol{p^\prime}, g)=d(\boldsymbol{p^\prime}, g)$, 
$\boldsymbol{p^\prime} = \{p_i^\prime\}_{i=1}^n$ is the joint next-step positions predicted by $n$ agents.
% Imagined next position $p_{t}^{k\prime}$ is utilized for the calculation of the task-specific reward as in Section \ref{subsec:4.1} for imagination policy. 
For example, in the shape formation task of snake robots, we use the Wasserstein distance \cite{villani2009optimal} between the imagined next-step position $\boldsymbol{p^\prime}$ and goal positions $g$ as the inverse of the reward for imagination policy, 
% \begin{equation}
$\displaystyle W(\boldsymbol{p^\prime}, g )= \inf_{\pi\in \mathcal{T}(\boldsymbol{p^\prime}, g)}\int_{\mathbb{R}^3 \times \mathbb{R}^3}|x-y|d\pi(x, y)$.
% \label{equ: w_distance}
% \end{equation}
% where $\boldsymbol{p^\prime_{t}} = \{p_{t}^{k\prime}\}_{k=1}^{n}$ describe the distribution of the anticipated snake shape, $\boldsymbol{p_g} = \{p_g^{k}\}_{k=1}^{n}, p_g^{k}\in \mathbb{R}^3$ denotes the target snake shape.
The goal of the imagination policy is to maximize the accumulated reward $J(\psi)=\mathbb{E}_{\boldsymbol{p_t^{\prime}},s_t}\left[\sum_t \gamma^t R^H(\boldsymbol{p_t^{\prime }}, g)\right]$.

The modular torque control policy receives rewards by both following the imagined displacement as prescribed by the imagination policy and completing the task. Agent $i$ executes action $a_{i,t}$ induced by the torque control policy $\pi_\theta(.|o_{i,t}, g)$, and reaches position $p_{i,t+1}\in \mathbb{R}^3$. The ``follow-the-imagination'' reward $r_f$ is defined as the sum of errors between actual position $p_{i,t+1}$ agent $i$ reaches and the imagined next-step position $p_{i,t}^{\prime}$.
\begin{equation}
    r_f = \sum_{i=1}^n|p_{i,t+1}- p_{i,t}^{\prime}|
\end{equation}
The reward function for the torque control policy can be described as $R = r_t + r_f$, where $r_t$ is a task-specific reward as shown in Table \ref{tab: r_t} in Section \ref{subsec: env}.

The shared imagination policy and the shared torque control policy are jointly trained following the framework of Multi-Agent PPO (MAPPO)\cite{yu2022surprising}.  
% In long-horizon tasks, the imagination policy helps to break down a distant goal into more attainable subgoals, providing insights for torque control policy to infer the low-level torque commands. Given a well-conceived imagination policy in place, the low-level policy can be guided more effectively toward a reasonable intermediate state and avoids getting stuck in a local optimal.
For each policy, we train two separate neural networks, the actor network and the critic network, $\pi_\theta(.|o_i, g)$ and $V_\phi(s)$ for the torque control policy, and $\pi^H_\psi(.|o_i, g)$ and $V^H_\omega(s)$ for the imagination policy, respectively. 
% Actor network for control policy $\pi_{\theta}$, maps agent observations $o_i$ and to the mean and standard deviation vectors of a Multivariate Gaussian Distribution, from which an action $a_i$ is sampled, in continuous action spaces. Similarly, the actor network for imagination policy $\pi^H_\psi$ outputs the distribution of the imagined per-step displacements, from which $p_{t}^{k\prime}$ is sampled. 
The critic networks, $V_{\phi}$ and $V^H_\omega(s)$, map the joint state to value: $S \rightarrow \mathbb{R}$.

The actor network for the torque control policy $\pi_{\theta}$ is trained to maximize the objective
\begin{equation*}
\begin{split}
    L(\theta)\hspace{-3pt} & = \frac{1}{Bn} \sum\limits_{k = 1}^{B} \sum\limits_{i = 1}^n \text{min}( r_{\theta, i}^{(k)}A_i^{(k)}, \text{clip}( r_{\theta, i}^{(k)}, 1 - \epsilon, 1 + \epsilon) A_i^{(k)}) \\
    & + \sigma \frac{1}{Bn} \sum\limits_{i = 1}^{B} \sum\limits_{k = 1}^n \hspace{-2pt} H[\pi_{\theta}(o_i^{(k)}))]
\end{split}
\label{equ-actoropt}
\end{equation*} 
where $\displaystyle{r_{\theta, i}^{(k)} = \frac{\pi_{\theta}(a_i^{(k)}|o_i^{(k)})}{\pi_{\theta_{old}}(a_i^{(k)}|o_i^{(k)})}}$, $A_i^{(k)}$ is the estimated advantage computed using the Generalizable Advantage Estimation (GAE)\cite{schulman2015high}, $H$ is the policy entropy, and $\sigma$ is the entropy coefficient hyperparameter. 
The critic network $V_\phi$ is trained to minimize the loss function
\begin{equation*}
\begin{split} 
L(\phi) &= \frac{1}{B}\sum\limits_{k = 1}^{B} \text{max}\lbrack(\text{clip}(V_\phi(s^{(k)}),V_{\phi_{old}}(s^{(k)})-\varepsilon,\\ 
&V_{\phi_{old}}(s^{(k)})+\varepsilon)-\hat{R})^2, (V_\phi(s^{(k)})-\hat{R})^2\rbrack
\end{split}
\label{equ-criticopt}
\end{equation*}
where $\hat{R}$ is the discounted reward-to-go, $B$ refers to the batch size, and $n$ refers to the number of agents. The actor network and critic network for the imagination policy, $\pi^H_\psi$
and $V^H_\omega$ are optimized similarly.

\subsection{Self-attention Mechanism}
\label{subsec: attention}
The coordinated behavior of a snake robot is heavily dependent on the cooperation of its highly coupled modules. 
We incorporate the self-attention \cite{vaswani2017attention} mechanism into the multi-agent training of the shared control policy, aiming to provide global information for each agent by augmenting inter-module communication. 
% Experiments in section \ref{sec:ablation} validated the effectiveness of this attention mechanism to enhance better coordination. It is demonstrated that this attention mechanism can successfully capture meaningful global information for each agent.

The self-attention encoder takes joint observation sequence $\boldsymbol{o} = \{o_i\}_{i=1}^n \in \mathbb{R}^{n\times d}$ as input, where $n$ is the sequence length, and $d$ is the observation dimension.  The self-attention encoder maps $\boldsymbol{o}$ to new latent space observation sequence 
% $\boldsymbol{o^h} = \{o_i^h\}_{i=1}^n$  
of equal length $n$ as shown in Fig. \ref{fig: overview}. Attention is calculated as:
\begin{equation}
    {\rm Attention}(Q,K,V)={\rm softmax}(\frac{QK^T}{\sqrt{d_k}})V
\end{equation}
Where $Q,K,V$ are query, key, and value, $Q=XW_q\in\mathbb{R}^{n\times r}, K=XW_k\in\mathbb{R}^{n\times r}, V=XW_v\in\mathbb{R}^{n\times f}$, $W_q, W_k\in\mathbb{R}^{d\times r}, W_v\in\mathbb{R}^{d\times f}$ are linear transformations. 
% Each layer contains a multi-head attention sublayer and a feedforward sublayer, with residual connection in between as in \cite{vaswani2017attention}. 
By calculating the attention, the new local observation for agent $i$ incorporates global information in the form of the weighted sum of $n$ transformed local observations of $n$ agents.

% \subsection{Algorithm Summary}
% \label{subsec:algo}
% The full algorithm is summarized in Algorithm \ref{algo:algo}. $\pi_\phi$, $V_\theta$, $\pi^H_\psi$, $V^H_\omega$ are jointly optimized using Adam optimizer.
\begin{algorithm}[t]
\caption{Training of the control and imagination policy}
\label{algo:algo}
\begin{algorithmic}[1]
    \STATE \textbf{Notation:} number of agents $n$, number of episodes $N_e$, episode length $T$
    \\\hrulefill
    % \STATE  Initialize $\pi_\theta$, $V_\phi$, $\pi^H_\psi$, $V^H_\omega$ \; 
    \FOR{$e$  \textbf{in} 1 \textbf{to} $N_e$}
    \STATE data buffer $D \leftarrow \{\}$
    % \FOR{$i = 1$ {\bfseries to} $batch\_size$}
    % \STATE $\tau = []$ empty list
    % \STATE initialize actor, critic RNN states
    % , $h_{0, V}^{(1)}, \dots h_{0, V}^{(n)}$ , $h_{0, \pi}^{(1)}, \dots h_{0, \pi}^{(n)}$
    \FOR{$t$ \textbf{in} 1 {\bfseries to} $T$}
    \FOR{$i$ \textbf{in} 1 to $n$}
    \STATE sample $a_{i,t} \sim \pi(o_{i,t}, g ; \theta)$
    \STATE sample $\Delta p_{i,t} \sim \pi^H(o_{i,t}, g; \theta)$
    \STATE $v_{i,t} \leftarrow V(s_t, g; \phi)$
    \STATE $v_{i,t}^H\leftarrow V^H(s_t, g; \phi) $
    \ENDFOR 
    \STATE Execute $\boldsymbol{a_t}$, %actions % $\boldsymbol{a_t}$
    $s_{t+1} \sim P(\cdot | s_t, \boldsymbol{a_t})$, predict $\boldsymbol{\Delta p_{t}}$, receive rewards $r_t, r^H_t$, observe $\boldsymbol{o_{t+1}}$ from $s_{t+1}$
    \STATE $D \leftarrow D \cup \{ ( s_t, \boldsymbol{o_t}, \boldsymbol{a_t}, \boldsymbol{\Delta p_{t}}, r_t, r^H_t, s_{t+1}, \boldsymbol{o_{t+1}} ) \} $
    \ENDFOR
    \STATE Compute advantage $\hat{A}$ and reward-to-go $\hat{R}$ on $D$
    % \STATE Compute reward-to-go $\hat{R}$ on $\tau$ and normalize with PopArt
    % \STATE Split $\tau$ into chunks of length $L$, append to buffer $D$
    % \FOR{l = 0, 1, .., T//L}
    % \STATE $D = D \cup (\tau[l : l + T, \hat{A}[l : l + L], \hat{R}[l : l + L])$
    % \ENDFOR 
    % \ENDFOR 
    % \FOR{mini-batch $k=1,\dots,K$}
    % \STATE $b \leftarrow$ random mini-batch from D with all agent data
    % \FOR{each data chunk $c$ in the mini-batch $b$}
    % \STATE update RNN hidden states for $\pi$ and $V$ from first hidden state in data chunk
    % \ENDFOR
    % \ENDFOR
    \STATE Adam update $\theta$, $\phi$, $\psi$, $\omega$ with $D$
    % \ENDFOR 
    \ENDFOR
\end{algorithmic}
\end{algorithm}

% We approximate the actor $\pi_\phi$, critic $V_\theta$, imagination policy $\pi^H_\psi$, imagination critic $V^H_\omega$
% with neural networks parametrized by $\phi$, $\theta$, $\psi$ and  $\omega$ respectively, and train them jointly using Adam optimizer. As the imagination policy $\pi^H_\psi$ learns to predict the next-step shape, it offers more relevant supervision to guide the behavior of the control policy, allowing more complex tasks. 

\section{Experiments}

The design of our COMPOSER framework is motivated by three hypotheses.
First, compared to centralized policies, the modular policy employed by COMPOSER can improve learning efficiency.
Second, the modular policy used by COMPOSER provides policy-level robustness against individual actuator malfunction.
Third, by using modular policy, COMPOSER generalizes better to snake robots with different numbers of joints.
To show this, we conducted extensive experiments across five snake robot tasks, evaluating success rates, robustness, and zero-shot generalizability to validate our hypotheses. We further illustrate the contributions of the imagination policy and the self-attention mechanism through ablation studies and visualization.

% In this section, we first introduce five snake robot tasks and five baselines in Section \ref{subsec: env}. In Section \ref{sub_sec: results}, we address three questions: Can our proposed modular policy COMPOSER learn successful control policy? Is COMPOSER more robust against snake robot individual actuator malfunction? Does COMPOSER exhibit better generalizability to unseen snake robots with different joint numbers? We further validate and visualize the effectiveness of the attention and imagination policy in Section \ref{sec:ablation} and Section \ref{sec:visualization}. 

\subsection{Experiment Setup}
\label{subsec: env}

\begin{figure} [!ht]
\centerline{
\includegraphics[width=0.5\textwidth, page=12, trim = 0.0cm 5.7cm 4.1cm 0cm, clip]{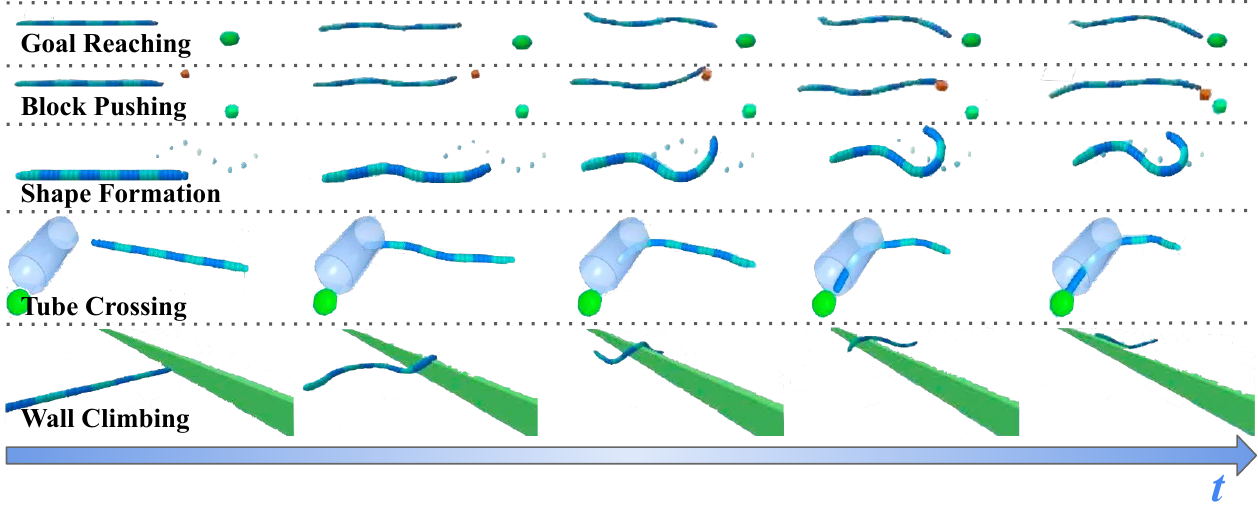}}
\vspace{-2.5ex}
\caption{Snapshots of COMPOSER across five snake robot tasks. 
% From top to bottom, we introduce five snake robot tasks, goal-reaching, block pushing, shape formation, tube crossing, and wall climbing. 
% Tasks are designed to showcase the structural advantage of snake robots compared with other types of rigid robots. In the goal-reaching and wall-climbing tasks, the snake robot exhibits a 3D locomotion ability akin to its biological counterpart. In the tube crossing and wall climbing tasks, the snake robot showcases adaptive and compliant interaction with the environment. Shape formation tasks emphasize the robot's flexibility and deformability, while the block-pushing task highlights the snake robot's precise manipulation capability.
}
\vspace{1ex}
\label{fig_env}
\end{figure}

\textbf{Environments.} We introduce five snake robot tasks, Goal Reaching, Block Pushing, Shape Formation, Tube Crossing, and Wall Climbing as shown in Fig. \ref{fig_env}. Our simulation environment is adapted from SomoGym ~\cite{graule2022somogym}.
% We examine our proposed imagination-guided self-attention modular controller on these five tasks. 
% Each experiment is run with four seeds to report the mean and the standard error. 
The reward is calculated as the sum of instant rewards across an episode for each environment. Once the robot completes the task, which is determined by the task-specific success criterion as shown in Table \ref{tab: r_t}, or the maximum episode length is reached, the episode terminates.

\begin{table}[t]
\caption{Task reward $r_t$ and success critrrion}
\vspace{-2ex}
\centering
\scalebox{0.8}{
\begin{tabular}{ccc}
\toprule
Task            & Reward $r_t$                              & Success criterion   
\\
\midrule
Goal Reaching   & $-d(head, goal)$                           & $d(head, goal)<d_g $                       \\ \midrule
Block Pushing   & $-d(head, object) - d(object, goal)$       & $d(object, goal)<d_g $                     \\ \midrule
Shape Formation & $-\sum_i^n d(agent_i, goal_i)$    & $\sum_i^n d(agent_i, goal_i)<d_g$ \\ \midrule
Tube Crossing   & $-d(head, tube) - d(head, goal)$           & $d(head, goal)<d_g  $                      \\ \midrule
Wall Climbing   & head position & snake tail over the wall        \\
\bottomrule                    
\end{tabular}}
\label{tab: r_t}
\end{table}

% The snake robot in the simulation is approximated by serially linked
% rigid segments connected by spring-loaded joints as in \cite{graule2022somogym}. Actuators are evenly distributed along the snake robot, and controlled via torque control. The actuators are equipped with two independently actuated bending axes, which produce anisotropic friction with the ground plane. 
The action space contains normalized torques for each actively controlled
degree of freedom (DOF) of each actuator. The action  $ a_i$  is represented by two scalar $[a_1, a_2]$ normalized to $[-1, 1]$, corresponding to two independently actuated bending axes in one actuator. 
In the multi-agent setting, each agent takes action only based on its own observation including local positions, velocities, orientations, and applied torques. For monolithic baseline training, the control policy takes global observations as input and outputs the action including commands for $n$ agents. 
\textbf{Baselines.} We compare our method with five baselines to investigate the effectiveness of COMPOSER. We apply the same hyper-parameters of baseline algorithms from their original papers.
PPO is a strong on-policy RL method. 
We use PPO with zero padding as a monolithic trained baseline to learn a centralized policy.
% that is widely used especially in environments with continuous action spaces. 
Following the setup in ~\cite{chen2018hardware}, we zero-pad the states and actions to the maximum dimension across all agents with different lengths.
MAT  is an on-policy few-shot learner on unseen tasks regardless of changes in the number of agents~\cite{wen2022multi}.
MAT\_dec  is a CTDE-variant of MAT~\cite{wen2022multi}.
% for a fair comparison, with a fully decentralized actor for each individual agent
SMP is a modular policy to generate locomotion behaviors for different skeletal structures via message passing~\cite{huang2020one}.
MAPPO is a strong multi-agent reinforcement learning baseline with competitive sample-efficiency~\cite{yu2022surprising}.

% Control effort: we define control effort as the sum of joint torques of all agents throughout each episode. We evaluate the average control effort in 100/1000 episodes. 

% Smoothness: To measure the continuity of the trajectory, we propose to compare the smoothness. The joint torque is normalized by making all joints start at -1 and
% end at 1 in all trajectories. Smoothness is defined as the sum of the L1 distances of the position, velocity, and joint torque between neighboring frames.

% The first metric mainly reflects whether the learned policy can complete the task. Control effort and smoothness can indicate whether the multi-agent system coordinates efficiently and smoothly.

\subsection{Results Comparison and Analysis}
\label{sub_sec: results}

We evaluate our COMPOSER framework in the following three aspects: learning efficiency in terms of task success rate, robustness against hardware failure, and zero-shot generalization to previously unseen robots.

\begin{comment}

To address the above three questions:
\begin{itemize}
    \item We first compare the success rate of COMPOSER across 5  tasks with 5 baselines.
    \item Next we examine the robustness of COMPOSER against hardware failures including sensor and actuator malfunction.
    \item Then we examine the zero-shot generalizability by applying the trained policy on unseen robots with different joint numbers.
\end{itemize} 
\end{comment}

% We answer the above questions by conducting experiments in five snake robot tasks. 
% Appendix materials contain more experiment details.

\textbf{Success Rate.}
\label{sec: general performance}
% trim=left bottom right top
\begin{figure*} [htbp]
\includegraphics[width=1\textwidth, page=1, trim = 0.cm 9.5cm 0.cm 6.2cm, clip]{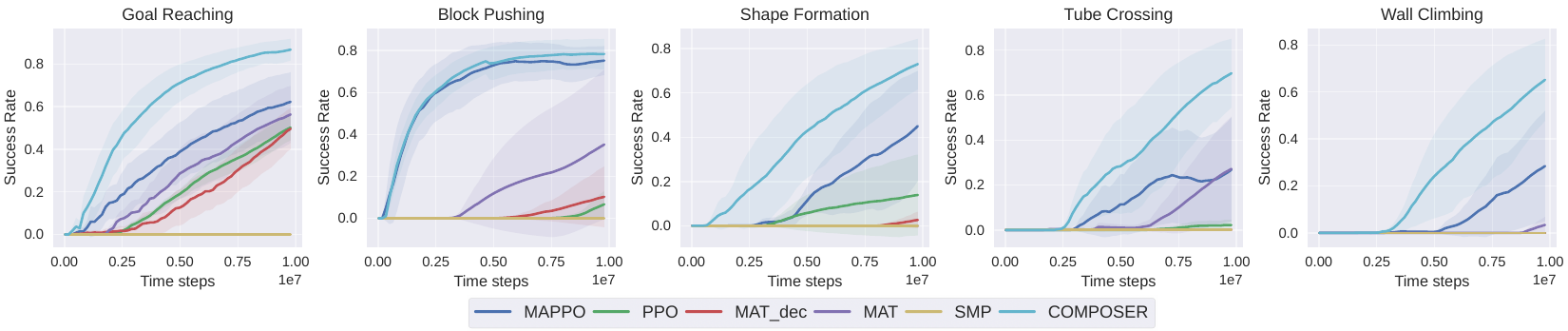}
\vspace{-4.5ex}
\caption{Learning curves of different methods in the tasks of (a) \textit{Goal Reaching}, (b) \textit{Block Pushing}, (c) \textit{Wall Climbing}, (d) \textit{Shape Formation}, and (e) \textit{Tunnel Crossing}. The evaluation rewards under desired goals are averaged over three seeds, and the shaded region represents the standard error.
}
% \vspace{-5pt}
\label{fig:exp_1}
\end{figure*}
In Fig. \ref{fig:exp_1}, we illustrate the performance of our proposed method and the baselines across five snake robot tasks. The results are averaged over three seeds, and the standard errors are provided. In almost all tasks, COMPOSER outperforms monolithic trained zero-padding PPO and other multi-agent RL baselines including MAPPO, MAT, MAT\_dec, and SMP. Monolithic trained PPO shows clear defects indicating the difficulties inherent in exploring a high-dimensional action space for snake robots. This observation validates the effectiveness of modular policy to tackle this high-dimensional challenge with decentralized control.  
For modular policy baselines, MAPPO is the strongest one while our method still outperforms it by a large margin, in terms of lower success rates and sample efficiency, showing the effectiveness of the self-attention mechanism and imagination policy.
% a shared modular policy trained among agents that might not necessarily be homogeneous—yet still achieves surprising cooperation without relying on communication. 
MAT features a complex model structure with both an encoder and a decoder, resulting in slow learning growth compared to COMPOSER.
MAT\_dec, being fully decentralized without the encoder for global observation embedding and the decoder that grants agents access to preceding agents' actions in MAT, significantly underperformed in comparison to our proposed method.
SMP does not manage to learn a successful control policy across all five tasks, 
% potentially influenced by the fact that the graph representation of the snake robot's morphological structure is simply a line. 
% Another possible reason could be that SMP is Q-learning based, which might not be well-suited for tasks involving continuous spaces, 
which aligns with results reported in ~\cite{graule2022somogym}, that TD3 has also shown no relevant improvement in nearly all training.% The corresponding model is then evaluated on 100 episodes, with randomly sampled target position in \textit{Reach} task. The evaluation results including episode length and success rate are shown in Table \ref{tab:episode_length}. The upper arrow means higher is better, and the down arrow means lower is better. The best performance of each task is marked in bold.
% \begin{table}[htbp]
% \caption{Performance evaluations of success rate and standard deviation}
% \centering
% \begin{tabular}{ccccccc}
% \toprule
%  & Ours & MAPPO & MAT & SMP & Zero-padding PPO \\
% \midrule
% Reach &      &       &     &     &                  \\
% Push  &      &       &     &     &                  \\
% Shape &      &       &     &     &                  \\
% Tube  &      &       &     &     &                  \\
% Wall  &      &       &     &     &                   \\
% \bottomrule

% \end{tabular}
% \label{tab:episode_length}
% \end{table}
% \begin{figure}[htbp]
%     \centering
%     \includegraphics[width=0.40\textwidth,height=2in]{figures/exp_2.png}
% \caption{Episode length of different methods in the task of \textit{Tunnel crossing}. 
% }
% \label{fig:exp_2}
% \end{figure}

\textbf{Robustness against Hardware Failures. }
\label{sec: robustness}
\begin{table*}[!]
\centering
\caption{Evaluations of robustness against hardware failures with different corruptions in eight-agent snake \textit{Goal Reaching} tasks}
\vspace{-2ex}
\label{tab: robustness}
\resizebox{1\textwidth}{!}{%
\begin{tabular}{@{}lcccccccccccc@{}}
\toprule
\multirow{2}{*}{Method} & \multicolumn{3}{c|}{0 fault} & \multicolumn{3}{c|}{1 fault action}& \multicolumn{3}{c|}{2 fault action} & \multicolumn{3}{c}{1 fault observation} \\ 
\cmidrule(l){2-13} 
 & \multicolumn{1}{c}{Success Rate $\uparrow$} & \multicolumn{1}{c}{Distance $\downarrow$} & \multicolumn{1}{c|}{Step$\downarrow$} & \multicolumn{1}{c}{Success Rate$\uparrow$} & \multicolumn{1}{c}{Distance $\downarrow$}& \multicolumn{1}{c|}{Step$\downarrow$} & \multicolumn{1}{c}{Success Rate$\uparrow$} & \multicolumn{1}{c}{Distance $\downarrow$}& \multicolumn{1}{c|}{Step$\downarrow$} & \multicolumn{1}{c}{Success Rate$\uparrow$} & \multicolumn{1}{c}{Distance $\downarrow$}& \multicolumn{1}{c}{Step$\downarrow$}
 \\\midrule
\multirow{1}{*}{PPO} & $ 0.51\pm0.15 $ & $ 3.14\pm 0.37$ & $ 102.42\pm6.74 $  & $ 0.28\pm 0.084$ & $ 5.14\pm 0.61$ & $ 111.62\pm3.77$ & $ 0.17\pm 0.06$ & $6.50\pm 0.77$ & $ 116.67\pm 2.31$ & $0.35\pm0.085 $ & $ 4.66\pm 0.099$ & $ 108.94\pm 2.97$\\ \midrule
\multirow{1}{*}{MAPPO} & $ 0.76\pm0.08 $ & $ 2.35\pm 0.09$ & $ 93.01\pm3.45 $  & $ 0.52\pm 0.093$ & $ 3.76\pm 0.54$ & $ 103.13\pm3.81$ & $ 0.28\pm 0.07$ & $5.30\pm 0.79$ & $ 112.36\pm 3.98$ & $\textbf{0.44}\pm0.14 $ & $ 5.20\pm 1.26$ & $ 106.29\pm 5.03$\\ \midrule
% \multirow{1}{*}{MAT} & $ \pm $ & $ \pm $ & $ \pm $ & $ \pm $ & $ \pm$ & $ \pm $ & $ \pm $ & $\pm $ & $ \pm $ & $\pm $ \\ \midrule
% \multirow{1}{*}{SMP} & $ \pm $ & $ \pm $ & $ \pm $ & $ \pm $ & $ \pm$ & $ \pm $ & $ \pm $ & $\pm $ & $ \pm $ & $\pm $ \\ \midrule
% \multirow{1}{*}{MAPPO} & $ \pm $ & $ \pm $ & $ \pm $ & $ \pm $ & $ \pm$ & $ \pm $ & $ \pm $ & $\pm $ & $ \pm $ & $\pm $ \\ \midrule
\multirow{1}{*}{\textbf{COMPOSER (ours)}} & $ \textbf{0.90}\pm0.028 $ & $ \textbf{2.08}\pm 0.059$ & $ \textbf{82.78}\pm 1.94$ & $ \textbf{0.55}\pm  0.073$ & $ \textbf{3.67}\pm 0.35$ & $ \textbf{100.07}\pm  1.07$ & $ \textbf{0.33}\pm 0.058$ & $\textbf{5.19}\pm 0.38$ & $ \textbf{110.62}\pm 1.35$ & $0.39\pm0.033 $ & $ \textbf{4.61}\pm 0.37$ & $ \textbf{105.07}\pm 1.83$ \\
\bottomrule
\end{tabular}%
}
\vspace{-10pt}
\end{table*}
% Structural-level redundancy enables snakes to exhibit flexible and compliant behaviors. We show that the modular policy provides policy-level redundancy for the snake robot to better leverage its structural redundancy.
In Table \ref{tab: robustness},  we demonstrate the performance of different policies trained on a normally functioning snake robot for \textit{Goal Reaching} when tested in scenarios involving partial hardware failures. 
% where agents exhibit aberrant behaviors in \textit{Goal reaching}.
This partial hardware failure includes agent action corruption and agent observation corruption. 
In the case of action corruption, faulty agents are assigned actions with a value of zero. 
In the case of observation corruption, the observation of faulty agents is assigned a value of zero.
At the beginning of each episode, faulty agents are randomly selected based on the specified faulty probability, $p = n_{\rm fault}/n,$, where $n_{\rm fault}$ is the number of corrupted agents, and $n$ is the total agent number. 
Corrupted agents fail to follow the prescribed policy. 
% Faulty actions are assigned a value of zero, and we use zero in observation to indicate observation loss.
% These fault agents fail to follow the prescribed policy.
% and we assume in most general cases these individual failures go undetected within the multi-agent system. 

In Table \ref{tab: robustness}, we evaluate different policies on four scenarios, including zero corrupted agent, one corrupted action, two corrupted actions, and one corrupted observation.
The corresponding model is evaluated on 100 episodes, and results include success rate, final distance to the goal, and episode length.
The upper arrow means higher is better, and the down arrow means lower is better. The best performance of each task is marked in bold. 
We compare COMPOSER with monolithic trained baseline PPO and the strongest modular policy baseline MAPPO. COMPOSER demonstrates greater fault tolerance by having a shorter distance to the goal, fewer episode steps across all scenarios, and a higher success rate across action corruption scenarios, when compared to PPO and MAPPO, as shown in Table \ref{tab: robustness}.  
% COMPOSER not only exhibits superior performance across all settings but also showcases a lower degree of performance degradation when actuator fault probability goes higher as shown in Table \ref{tab: percentage}. However, COMPOSER experiences worse performance degradation in the presence of local observation loss when compared to PPO. 
In the case of one corrupted observation, MAPPO demonstrates a slightly higher success rate. This outcome could be attributed to the fact that the corruption of a single agent's local observation might lead to confusion in the observations of other agents, given the presence of the attention mechanism. In contrast, MAPPO is less affected by the corruption of local observations due to the lack of inter-agent communication.
% \begin{table}[h]
% \caption{Performance degradation under different fault probability}
% \centering
% \begin{tabular}{ccccc}
% \toprule
% & 1 fault action & 2 fault action & 1 fault observation  \\
% \midrule
% PPO &   54\%   & 33\%&      \textbf{68\%}                        \\
% COMPOSER  & \textbf{61\%}&    \textbf{36\%} &     43\%                         \\
% \bottomrule

% \end{tabular}
% \label{tab: percentage}
% \end{table}
\textbf{Zero-shot Generalizability to Snake Robots with Different Number of Joints. }
\label{sec: generalization}
% We show that the modular policy has the generalizability to match the flexibility and scalability of the continuum robot. 
To demonstrate zero-shot generalizability, we train the policy on a snake robot with 8 agents and directly apply the trained policy on previously unseen snake robots with 9 agents, 10 agents, and 11 agents. As shown in Table \ref{tab: generalization}, 
% the modular policy only trained on an 8 agent snake \textit{Goal reaching} task, can be generalized to 9 agent snake with $0.57$ success rate, which is $63\%$ of success rate for 8 agent snake. T
our modular policy demonstrates notable generalizability in comparison to PPO and MAPPO, particularly when applied to a 10-agent snake robot, where PPO and MAPPO almost always fail, but our proposed method has a 33\% success rate. This shows the potential of our method to be applied to a scalable modular snake robot.

% To demonstrate scalability, we train the policy on a continuum robot with 64 links and compare the performance with the policy trained under a centralized execution framework. as shown in Table \ref{tab:4-1}. TODO

\begin{table*}[!]
\centering
\caption{Evaluations of zero-shot generalization to snake robots with different lengths on \textit{Goal Reaching} task} 
\vspace{-2ex}
\label{tab: generalization}
\resizebox{\textwidth}{!}{%
\begin{tabular}{@{}ccccccccccccc@{}}
\toprule
\multirow{2}{*}{Method} & \multicolumn{3}{c|}{8 agent (32 links)} & \multicolumn{3}{c|}{9 agent (36 links)}& \multicolumn{3}{c|}{10 agent (40 links)} & \multicolumn{3}{c}{11 agent (44 links)} \\ 
\cmidrule(l){2-13} 
 & \multicolumn{1}{c}{Success rate $\uparrow$} & \multicolumn{1}{c}{Distance $\downarrow$} & \multicolumn{1}{c|}{Step$\downarrow$} & \multicolumn{1}{c}{Success rate$\uparrow$} & \multicolumn{1}{c}{Distance $\downarrow$}& \multicolumn{1}{c|}{Step$\downarrow$} & \multicolumn{1}{c}{Success rate$\uparrow$} & \multicolumn{1}{c}{Distance $\downarrow$}& \multicolumn{1}{c|}{Step$\downarrow$} & \multicolumn{1}{c}{Success rate$\uparrow$} & \multicolumn{1}{c}{Distance $\downarrow$}& \multicolumn{1}{c}{Step$\downarrow$}
 \\\midrule
\multirow{1}{*}{PPO} & $ 0.51\pm0.15 $ & $ 3.14\pm 0.37$ & $ 102.42\pm6.74 $  & $ 0.25\pm 0.01$ & $ 5.23\pm 0.26$ & $ 112.79\pm1.51$ & $ 0.067\pm 0.047$ & $7.10\pm 0.59$ & $ 120.43\pm 2.52$ & $0.033\pm0.023 $ & $ 8.51\pm 1.73$ & $ 122.13\pm 1.32$\\ \midrule
\multirow{1}{*}{MAPPO} & $ 0.76\pm0.08 $ & $ 2.35\pm 0.09$ & $ 93.01\pm3.45 $  & $ 0.47\pm 0.05$ & $ 3.61\pm 0.21$ & $ 107.99\pm1.58$ & $ 0.10\pm 0.07$ & $7.14\pm 1.42$ & $ 119.41\pm 2.68$ & $0.023\pm0.02 $ & $ 11.4\pm 1.70$ & $ 122.45\pm 0.71$\\ \midrule
\multirow{1}{*}{\textbf{COMPOSER}} & $ \textbf{0.90}\pm0.028 $ & $ \textbf{2.08}\pm 0.059$ & $ \textbf{82.78}\pm 1.94$ & $ \textbf{0.57}\pm  0.20$ & $ \textbf{2.83}\pm 0.40$ & $ \textbf{102.24}\pm  4.28$ & $ \textbf{0.35}\pm 0.18$ & $\textbf{4.20}\pm 1.49$ & $ \textbf{111.25}\pm 6.00$ & $\textbf{0.13}\pm0.09 $ & $ \textbf{7.26}\pm 2.72$ & $ \textbf{118.87}\pm 3.15$ \\
\bottomrule
\end{tabular}%
}
\vspace{-5pt}
\end{table*}

% \begin{table}[h]
% \caption{Performance evaluations of success rate and standard deviation}
% \centering
% \begin{tabular}{ccccccc}
% \toprule
%  & Ours & MAPPO & MAT & SMP & Zero-padding PPO \\
% \midrule
% reach &      &       &     &     &                  \\
% push  &      &       &     &     &                  \\
% shape &      &       &     &     &                  \\
% tube  &      &       &     &     &                  \\
% wall  &      &       &     &     &                   \\
% \bottomrule

% \end{tabular}
% \label{tab:4-2}
% \end{table}

\subsection{Ablation Study: Attention and Imagination}
\label{sec:ablation}
\begin{figure*} [htbp]
\includegraphics[width=1\textwidth, page=1, trim = 0.cm 9.5cm 0.cm 6.1cm, clip]{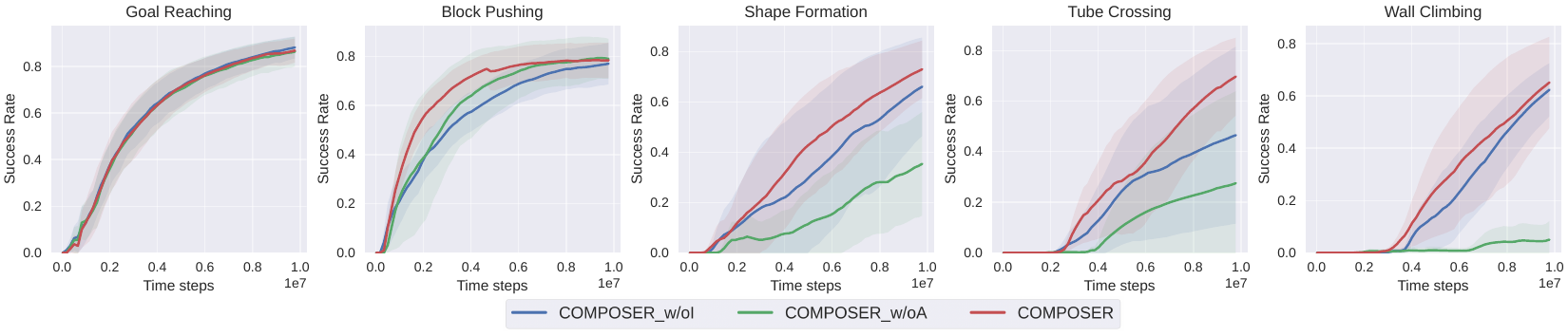}
\vspace{-4.5ex}
\caption{The comparison between COMPOSER and its ablations, COMPOSER without imagination policy (COMPOSER\_w/oI) and COMPOSER without self-attention mechanism (COMPOSER\_w/oA). From left to right, success rate learning curves are plotted for five tasks, (a) \textit{Goal Reaching}, (b) \textit{Block Pushing}, (c) \textit{Wall Climbing}, (d) \textit{Shape Formation}, and (e) \textit{Tunnel Crossing}. 
% For \textit{Goal reaching} and \textit{Block pushing}, COMPOSER\_w/oI and COMPOSER\_w/oA achieved similar performance. For the \textit{Shape formation} task, which demands not only locomotion but also desired deformation, and for \textit{Wall climbing} and \textit{Shape formation}, which are heavily dependent on effective interaction with the environment (tube and wall), the utilization of an imagination policy and self-attention mechanism demonstrates a clear advantage.
}
% \vspace{-10pt}
\label{fig:exp_ablation}
\end{figure*}

\begin{figure*} [htbp]
\includegraphics[width=1\textwidth, page=4, trim = 0.0cm 7.5cm 0cm 0cm, clip]{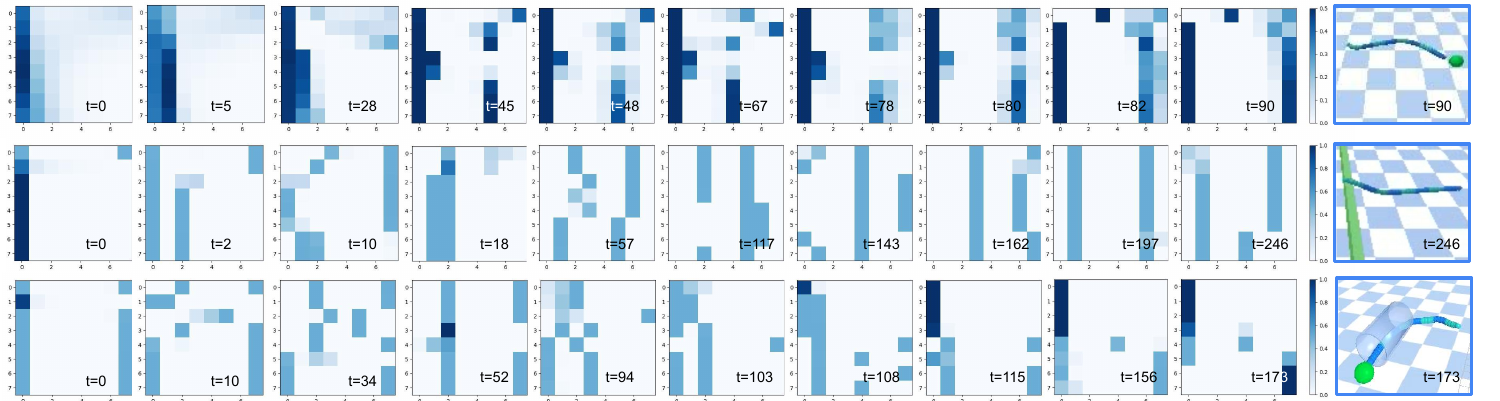}
\vspace{-4.5ex}
\caption{Visulaization of  attention matrixes $A\in\mathbb{R}^{8\times8}$ for an 8-agent snake. $A_{ij}$ is the attention score for agent $i$ attending to agent $j$. The higher the attention score is, the more agent $i$ attends to agent $j$. From left to right, we show transitions of the attention matrix throughout one successful episode for three tasks,  (a) \textit{Goal Reaching}, (b) \textit{Wall Climbing}, and (c) \textit{Tunnel Crossing}.
}
\label{fig:exp_attention}
\vspace{-10pt}
\end{figure*}

As shown in Fig. \ref{fig:exp_ablation}, we evaluate the performance of COMPOSER alongside its variants: one devoid of the imagination policy (COMPOSER\_w/oI) and another excluding the self-attention mechanism (COMPOSER\_w/oA) across five tasks. While COMPOSER\_w/oI and COMPOSER\_w/oA achieved a success rate similar to COMPOSER in simpler tasks such as \textit{Goal Reaching} and \textit{Block Pushing}, they fell notably short in tasks of increased complexity. In more challenging tasks, such as \textit{Shape formation} which requires desired deformation, and \textit{Wall climbing} and \textit{Shape Formation} which necessitate effective interaction with the environment, the ablations demonstrate markedly worse performance. COMPOSER enjoys a higher success rate by leveraging both the attention mechanism for enhanced coordination and the imagination policy for more efficient planning.

\subsection{Qualitative Visualization}
\label{sec:visualization}
In Fig. \ref{fig:exp_attention}, we visualize the heatmap of the attention matrix $A=\displaystyle{\rm softmax}(\frac{QK^T}{\sqrt{d_k}}), A \in\mathbb{R}^{8\times8},$ for an 8-agent snake. $A_{ij}$ is the attention score for agent $i$ attending to agent $j$. The first agent ($i=0$),  denotes the snake head, and the last agent ($i=7$) denotes the snake tail. 
As shown in Fig. \ref{fig:exp_attention}, our attention mechanism effectively captures different patterns for different tasks, and the attention dynamically moves to different body segments over time.
% implying that the attention mechanism effectively captured shared global information for individual agents. 
% This acquired information is subsequently combined with agents' local observations for decentralized action with a better understanding of the global state. 
This shows that our attention mechanism has the potential to encompass both task-related details and the underlying dynamics of the snake robot.
% The shared global information among all agents has the potential to encompass both task-related details and the underlying dynamics of the snake. 
In \textit{Goal Reaching} (the first row in Fig. \ref{fig:exp_attention}), the attention matrix consistently focuses on the snake head, which directly determines the shared reward and the success condition of the task.  
We also observe that a portion of the attention transitions from the snake head to the snake tail from $t=0$ to $t=90$, akin to the sinusoidal actuation pattern that generates locomotion through anisotropic friction \cite{hu2009mechanics}. 
In \textit{Wall Climbing} (the second row in Fig. \ref{fig:exp_attention}), the agents attend to different body segments that are interacting with the wall over time.
In \textit{Tunnel Crossing} (the third row in Fig. \ref{fig:exp_attention}), the attention is distributed over more segments than the first two tasks at the early stage, because its whole body has to adapt to the tunnel shape. Then, the agents attend more to the head part as the head of the snake gets close to its targeted position.
% Also, a similar observation can be made with the third row in Fig. \ref{fig:exp_attention}, where the attention shifts to the snake head after $t=108$, as the snake head tries to reach the target.
Notably, the colors across different agents are often nearly uniformly distributed within the same column of the heatmap, implying that $n$ agents achieve a shared consensus on global information through self-attention communication.

% \section{Coneection with Soft robot}
% It is worth noticing that the results achieved on snake robots can be extended to soft robots. Sanke robot as a continuum manipulator is a discretization of the soft robot, featuring a finite number of actuators, applying forces/torques to the backbone at a preselected set of locations. The snake robot studied in this paper is split into several discrete segments with revolute joints between them, and each segment is approximated by four rigid links, which is a Pseudo-Rigid Body (PRB) model for the soft robot manipulator. PRB theory has shown great potential in the description of a flexible body.  The PRB 3R model accurately describes the large deformation of the flexible body and has high computational efficiency. \cite{huang20193d}In the PRB approach, the backbone along a subsegment is modelled as a series of rigid links connected by springs\cite{su2009pseudorigid}. 
\section{Conclusion}
Controlling snake robots for dexterous and compliant behaviors is extremely challenging due to complex dynamics. 
To address this challenge, we introduce a scalable and robust modular policy that leverages the inherent modularity of snake robots. 
In this work, we formulate the snake robot control framework within a cooperative multi-agent reinforcement learning setting.  Extensive experiments are conducted across five tasks, showcasing the modular policy's notable efficiency in learning snake robot control, its robustness against hardware failures, and its generalizability for previously unseen robots.
Furthermore, we plan to extend the application of the trained modular policy to physical continuum robots and soft robots.

\bibliographystyle{IEEEtran}
\bibliography{ref}  % .bib

\end{document}